# AN ONTOLOGY CONSTRUCTION APPROACH FOR THE DOMAIN OF POULTRY SCIENCE USING PROTÉGÉ

P. Kalaivani[1] A. Anandaraj[2] K. Raja[3]

[2]Lecturer [3]Professor

[2&3]Department of Computer Science and Engineering, Narasu's Sarathy Institute of Technology. Salem,Tamilnadu,India

kalaipadman@gmail.com anandarajme@gmail.com raja_koth@yahoo.co.in

***Abstract***

The information retrieval systems that are present nowadays are mainly based on full text matching of keywords or topic based classification. This matching of keywords often returns a large number of irrelevant information and this does not meet the user's query requirement. In order to solve this problem and to enhance the search using semantic environment, a technique named ontology is implemented for the field of poultry in this paper. Ontology is an emerging technique in the current field of research in semantic environment. This paper constructs ontology using the tool named Protégé versioned 4.0 and this also generates Resource Description Framework (RDF) schemas and XML scripts for using poultry ontology in web.

## 1. INTRODUCTION

The recent growth in technology made information retrieval as an emerging technique. However, the current retrieval methods are essentially based on the string matching approach with the lack of semantic information and cannot understand the user's query requirement very well. Recall ratio and precision ratio can only be increased by semantic intelligence of retrieval systems. These deals with extending the existing web with conceptual metadata are more useful to machines.

Facet-based as well as traditional keyword based method retrieval operations are carried out based on the vocabulary. There is a certain degree of semantic missing. Ontology as a basis for the sharing of knowledge has been widely used in information science. Ontologies define domain concepts and the relationships between them, and thus provide a domain language that is meaningful to both humans and machines.

Ontologies are being defined for poultry science. The field poultry comes under veterinary science as animal husbandry. Poultry has two varieties of chickens. They are broadly classified into layers and broilers. The parents of layer chicks are White Leghorn and Rhode Island Red whereas White Cornish for the broiler birds. Layer birds are mainly meant for egg production and broiler birds are for meat production. The domains can further be classified into many classes and subclasses. The concepts from these ontologies can be used to annotate web resources. The web ontology language (OWL) is widely accepted as the standard language for sharing semantic web contents and is implemented through Protégé OWL.

Protégé is an ontology development environment with a large community of active users. Protégé has been extended with support for OWL, and has become one of the leading OWL tools.

The goal of this paper is to help poultry projects to get started with semantic web technology. This paper describes the classification of poultry science among the domains involved in it using protégé. This also defines the classes, properties and features such as reasoners to check semantic consistency. Finally, the paper shows the graphical view of the classes, generated RDF schemas and XML scripts which is used to link existing web resources into the semantic web.

## 2. ONTOLOGY CONSTRUCTION USING PROTÉGÉ

OWL ontology can be depicted as a network of classes, properties and individuals. Classes define names of the relevant domain concepts and their logical characteristics. Properties (also called as roles, attributes or slots) that defines relationships between classes, and allow assigning primitive values to instances. Individuals are instances of the classes with specific values for the properties [1].

The semantic web can be regarded as a network of ontologies and other web resources [1]. OWL ontology concepts can have references to concepts in other ontologies. The basic mechanism for this capability is ontology import (i.e., ontology can import resources from existing ontologies and create instances or specializations of their classes)

### *2.1 Classes*

The important view in the Protégé OWL plugin is the OWL classes. Classes describe concepts in the domain. This tab displays the tree of the ontology's classes on the left. The upper region of the class is shown in a form in the center. This form allows users to edit class metadata such as name, comments, and labels, in multiple languages. The widget in the right area of the form allows users to assign values for properties and description to a class.

Annotation properties can be used to add information (metadata-data about data) to classes. Ontologies can define their own annotation properties or reuse existing ones such as those from the Dublin core ontology [1]. In contrast to other properties, annotation properties do not have any formal meaning for external OWL components like reasoners, but they are an extremely important vehicle for maintaining project information. A typical use for annotation property in poultry field is to design concept that describes functionality of each class.

In this paper many classes and sub classes have been created under the field of poultry but due to lack of space only some of the classes are described elaborately.

Here, the class Health monitoring and disease control is a sub class of Breeder farm management and it has two sub classes namely bio security and vaccination. The class vaccination is again sub divided into a sub class prevention of diseases. The classification further proceeds to a large number of classes and sub classes. The representation of classes is depicted in Figure 1.

The editing of classes in carried out using the classes tab shown in Figure 2. The initial class hierarchy tree view should resemble the picture shown in Figure 2. The empty ontology contains one class called Thing. The class Thing is the class that represents the set containing all individuals. Because of this all classes are subclasses of Thing [2]. To add a class, the classes tab is selected, add subclass button is pressed. This creates a new class as a subclass of the selected class Thing.

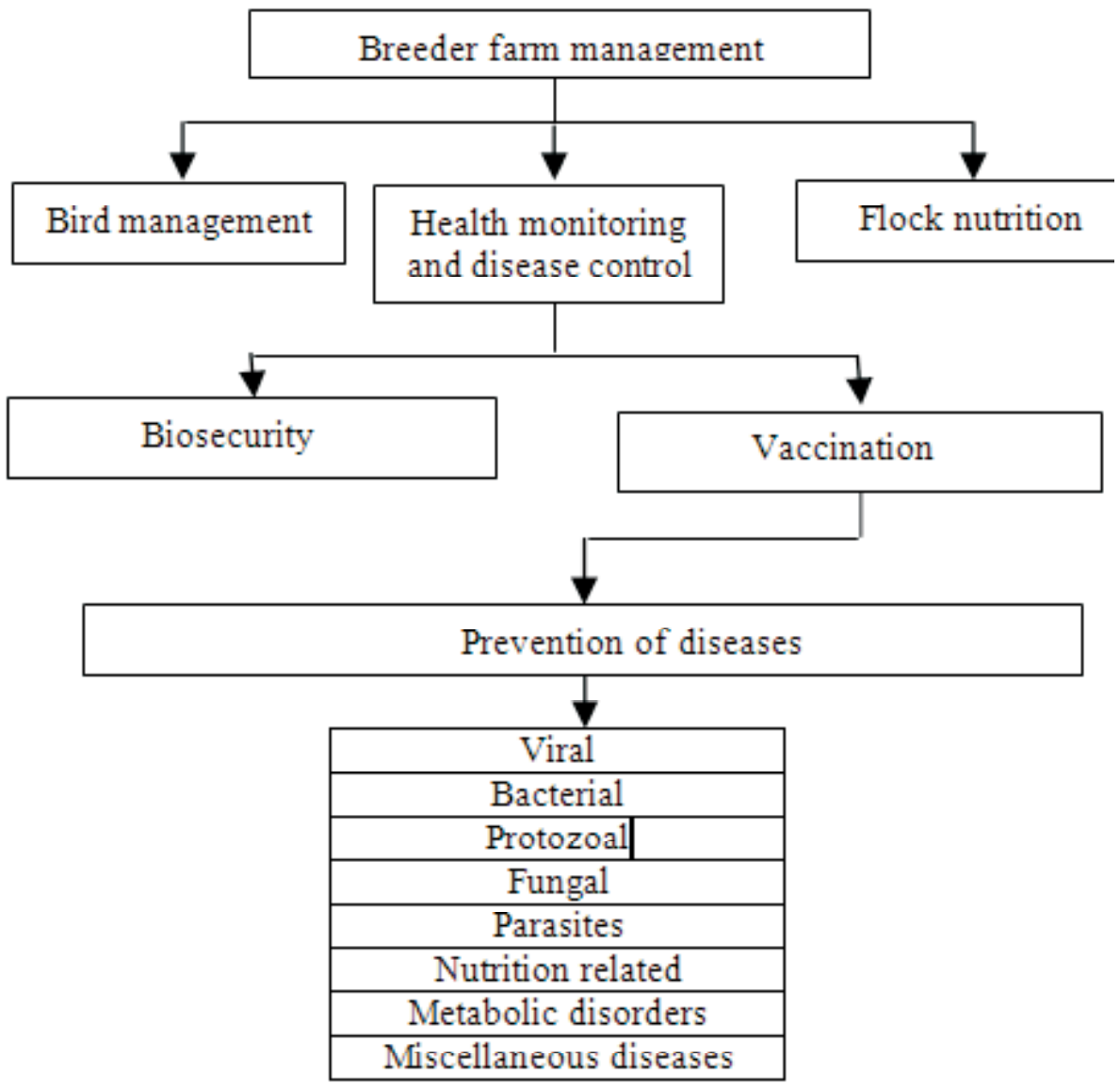

Figure 1: Representation of Classes

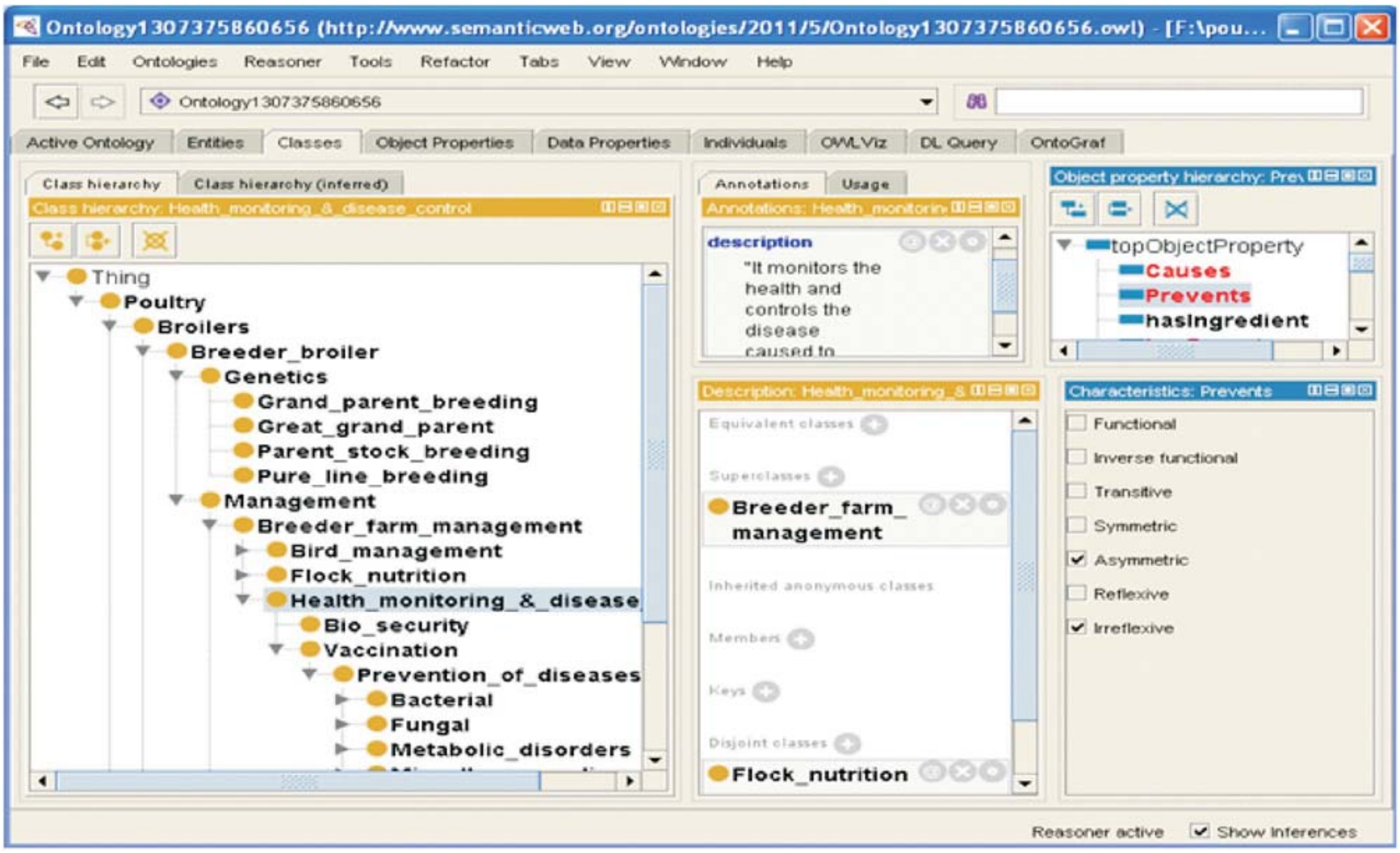

Figure 2: Description of classes with its views

### *2.2 Properties*

The properties widget of the OWL classes tab allows users to view and create relationships between classes. It provides access to those properties that could be used by the instances of the current class. The characteristics of a property are edited through the form shown in Figure 4. This form provides a metadata area in the upper part, displaying the property's name, annotations and so on, similar to the presentation in the class form.

There are two main types of properties viz. Object properties and Datatype properties. Object properties are relation between two individuals. Object properties link an individual to an individual whereas datatype property link an individual to an XML Schema Datatype value or an RDF literal (ie.They describe relationship between an individual and data values). OWL also has another property named annotation property, which is used to add information (ie.metadata - data about data) to classes, individuals and object/datatype properties.

This paper implements about twelve properties. Some of them are listed in the Table 1. Each of the property has its own characteristics and are described in Table 2. Each and every attribute of the property are independent and each has its own characteristics.

Table 1 : Properties used in Poultry

| S. No. | Name of the Properties |
|---|---|
| 1 | hasPeriod |
| 2 | IsPeriodOf |
| 3 | hasPreventivemeasure |
| 4 | isPreventivemeasureOf |
| 5 | Prevents |
| 6 | ispreventedBy |
| 7 | Causes |
| 8 | isCausedBy |

Table 2 : Characteristics of Properties

| S. No. | Name of the Characteristics |
|---|---|
| 1 | Functional |
| 2 | Inverse Functional |
| 3 | Transitive |
| 4 | Symmetric |
| 5 | Antisymmetric |
| 6 | Reflexive |
| 7 | Irreflexive |

This paper discusses about some of the object properties namely hasPreventive measure, isPreventivemeasureOf, Prevents, isPreventedBy, Causes, isCausedBy.

The class Health monitoring and disease control relates to the class Vaccination through the property hasPreventivemeasure and the class Vaccination is related to the class Health monitoring and disease control through the property isPreventivemeasureOf These two properties are inverses of each other. The classes Vaccination and Bacterial relates through Prevents and isPreventedBy property. The classes Bacterial and fowl typhoid are related through Causes and isCausedBy property. The characteristics of these properties are Antisymmetric and irreflexive.

The properties can be edited using the properties tab selecting either object properties or datatype properties. Annotations can also be added to the properties in order to describe about it. To create an object property switch to object properties tab, use the add object property button, this creates a new object property.

To set characteristics, the check box nearer to the property is checked. To specify properties to the classes domain and range are specified. In this case, the domain of Causes is Bacterial and its range is Fowl typhoid and the inverse of this will be for isCausedBy.

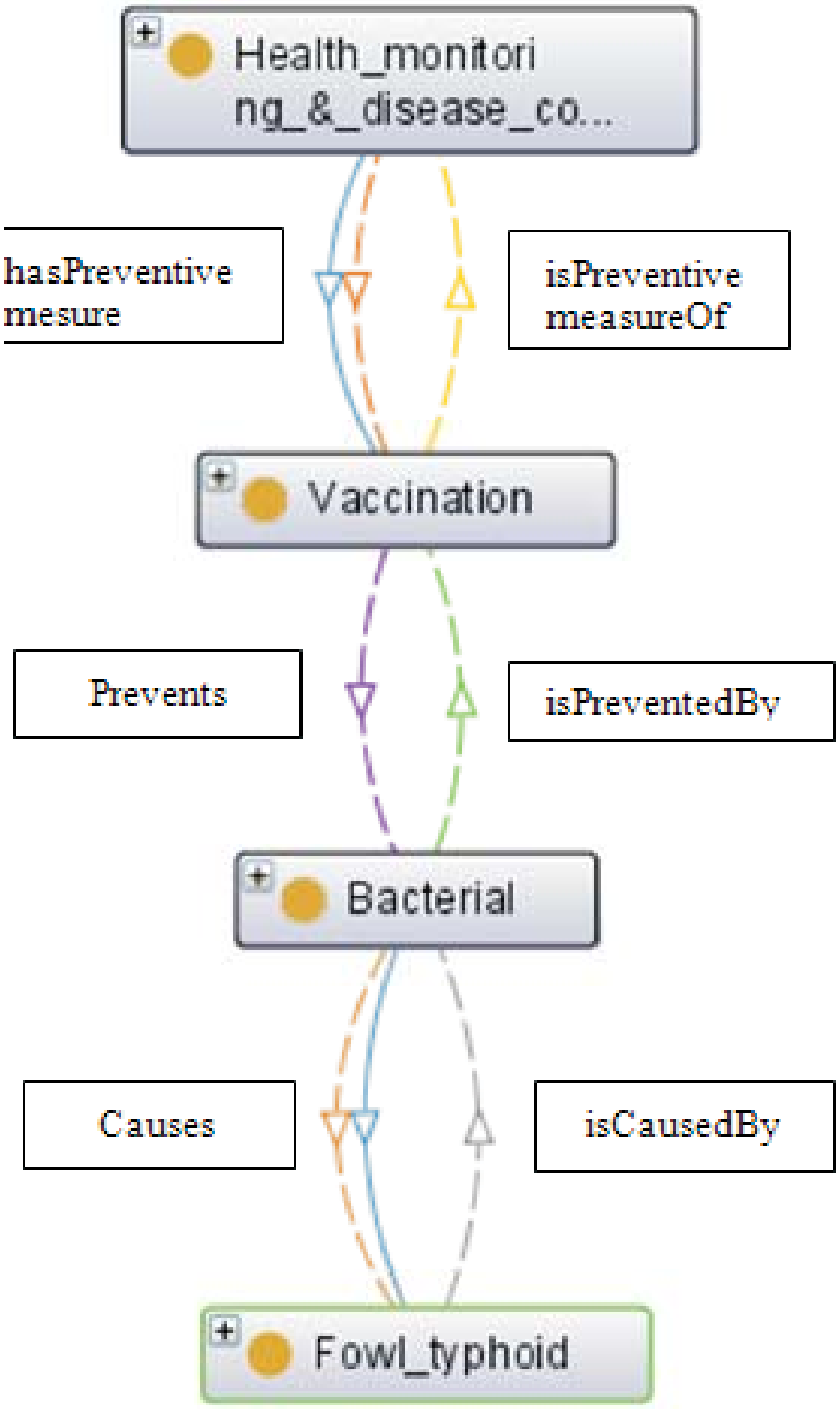

Figure 3: Property description using Classes

## 3. OPEN WORLD ASSUMPTION

The assumption is made by description logic, this denotes a lack of knowledge. The consequence is that if two classes Biosecurity and Vaccination are not defined as disjoint then it can have the individuals in common. The disjointness in classes plays a vital role in each of the class description. Creating a class and making it complement to another class is done here.

### *3.1 Using Reasoner*

The consistency can be checked through the reasoner. Protégé supports many reasoners. This paper uses Hermit as its reasoner. Reasoning means to infer new knowledge from the statements asserted by an ontology designer. The illegal mistakes committed by the developer are spotted out by the reasoner. The problem that is faced when the poultry ontology is developed is, due to a wrong setting of property characteristics, the reasoner displays error messages like inconsistent ontologies.

Reasoning capabilities are exploited to detect logical inconsistencies within the ontology. The error has been occurred while setting characteristics, asymmetric and reflexive to a same property. The consistency checks can help developer in an adequate manner while constructing the ontologies.

The important issue with reasoners is that OWL is not able to handle full expressivity. The specification distinguishes between OWL Full and OWL DL to indicate tractable language elements to reasoners. Ontologies which use metaclasses which is a OWL Full element cannot be classified. The conversion of OWL Full to OWL DL can be made using the classifier. Complete OWL Full syntax is not supported by protégé.

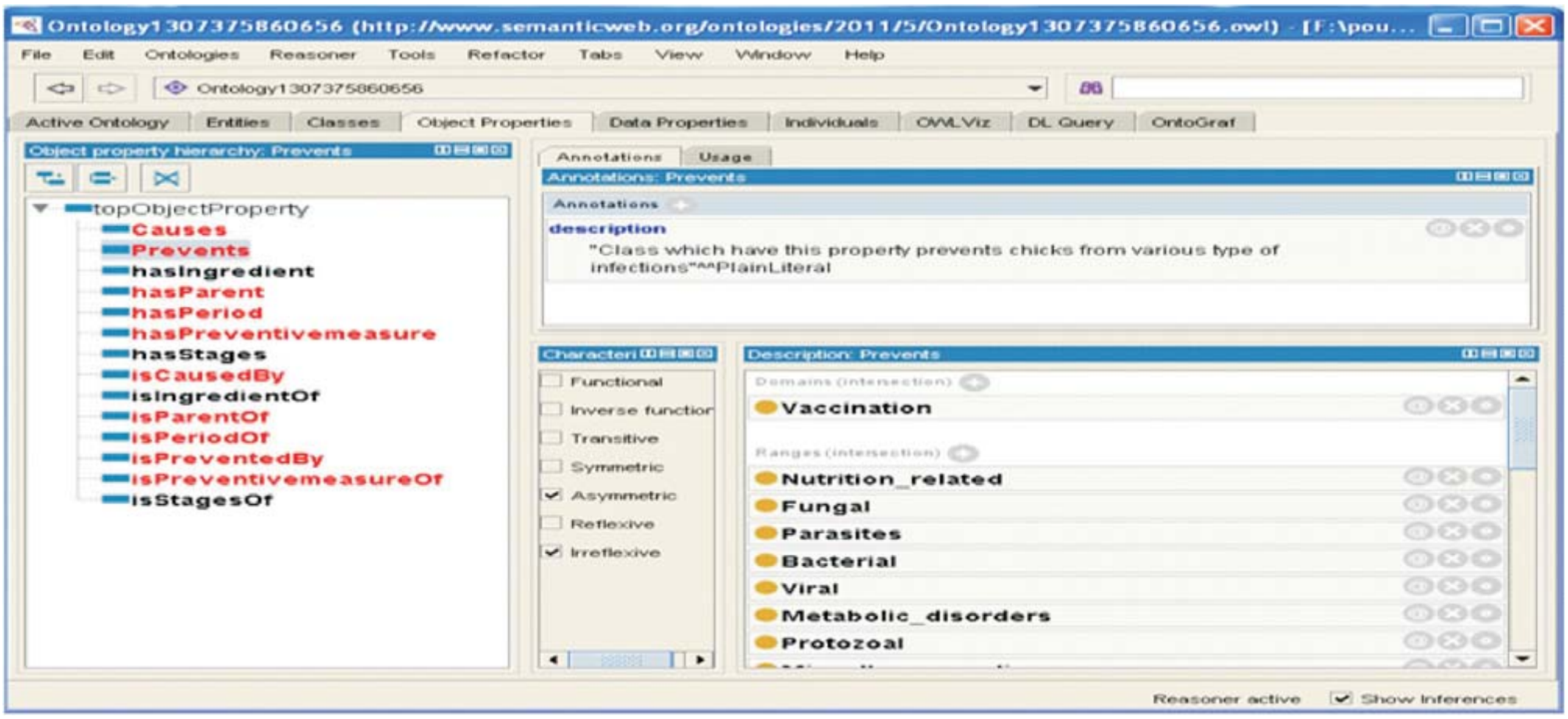

Figure 4: Object property for classes

### *3.2 RDF/XML Rendering*

The structure of any expression in RDF consists of triples, each consisting of a subject, a predicate and an object. A set of each triples is called an RDF graph. This can be illustrated using the node and arc diagram, in which each triple is represented as a node-arc-node link. In order to avoid conversion between different description languages, ontology needs a common language to express. XML has been used for this purpose since it has standards on data exchange. OWL ontology is most commonly serialized using RDF/XML syntax. The RDF/XML schemas are represented in Figure 5.

## 4. OVERALL VIEW OF POULTRY ONTOLOGY

Onto graphical view completely describes about each classes and sub classes that were created. Due to the lack of space only few classes and sub classes are mentioned. This graph also depicts the relationship that exists between each and every classes and sub classes. Different colors are used to distinguish between different properties.

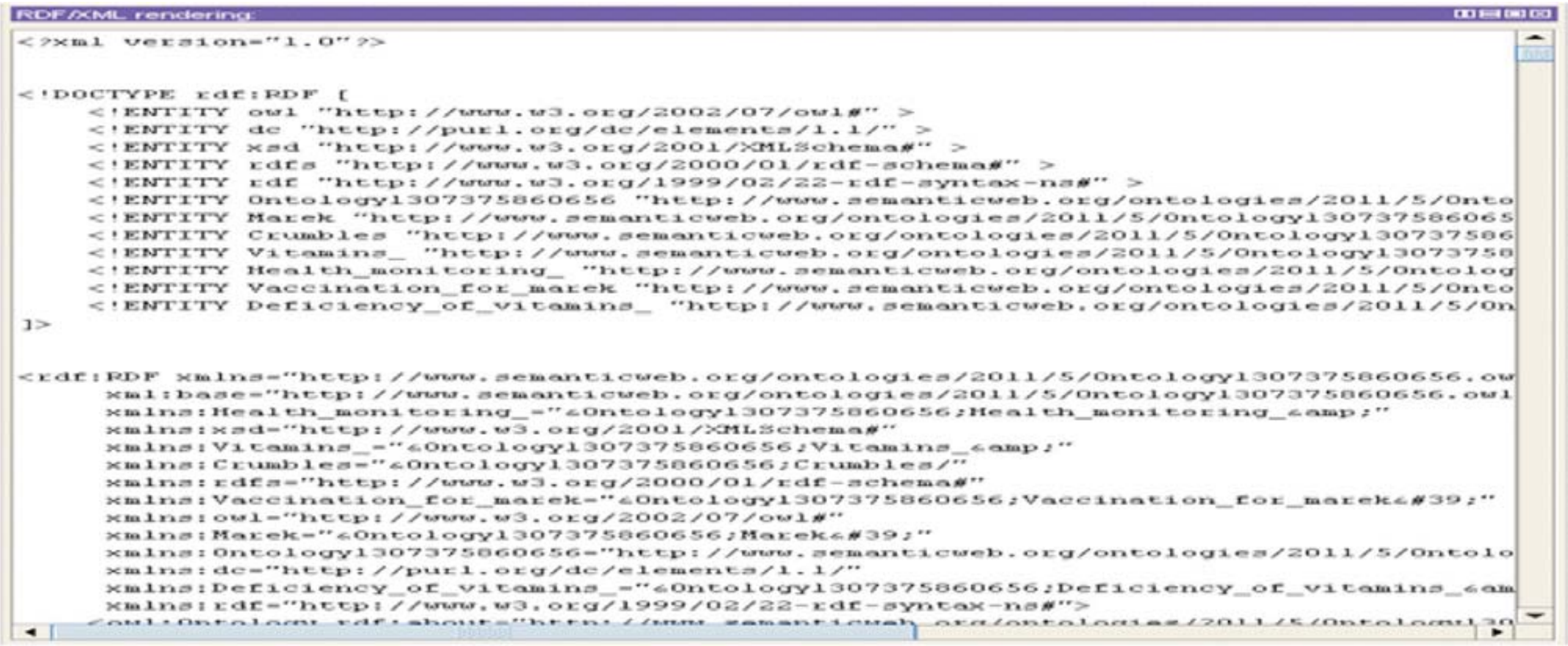

Figure 5: RDF/XML rendering

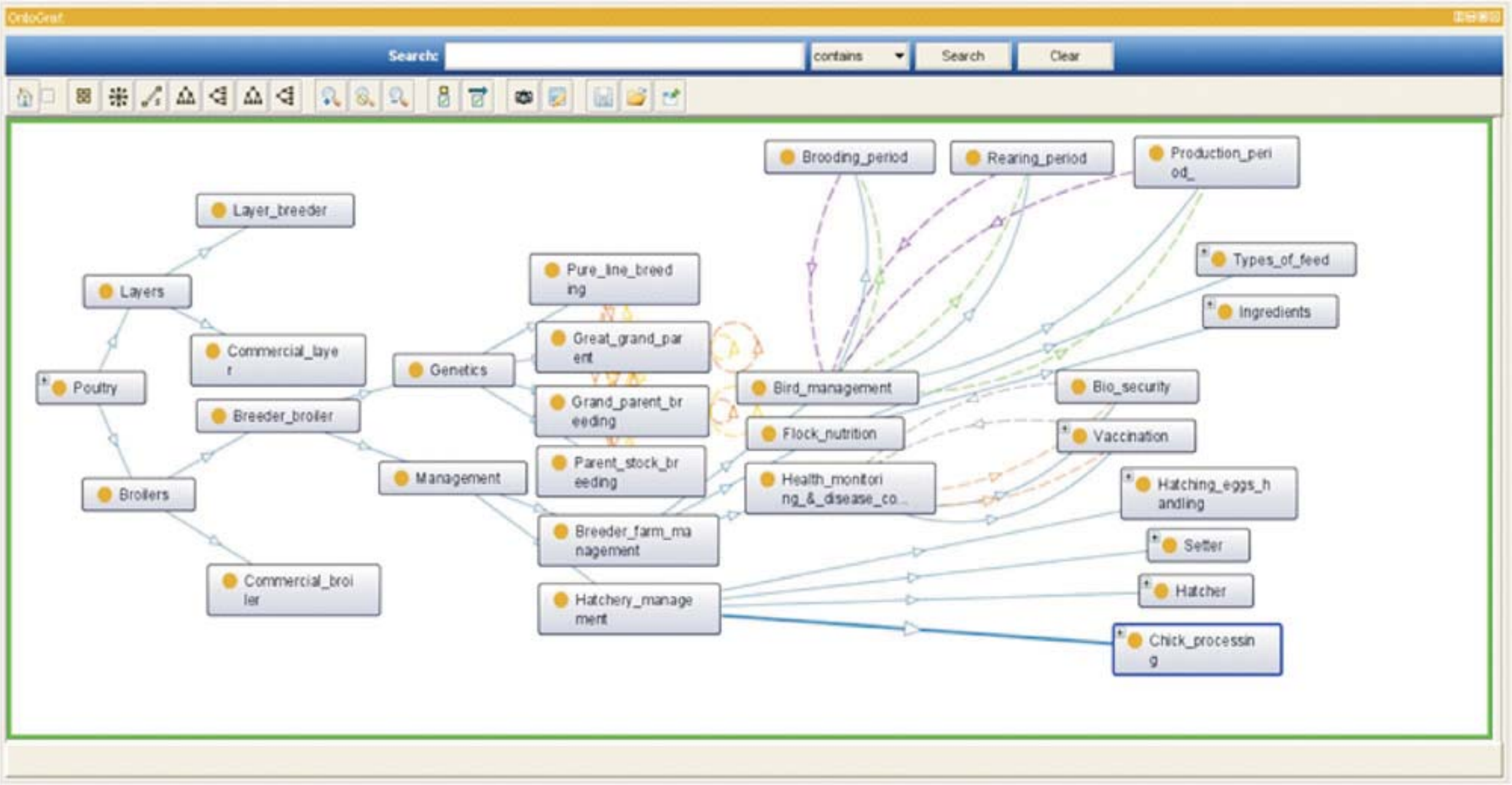

Figure 6: Poultry Ontology

## 5. CONCLUSION AND FUTURE WORK

This paper described a framework of an ontology construction for poultry domain to extract information about the field. Under this construction of framework of ontology, the doctors and other technicians who involved in the domain will achieve a mass of both the linguistic information and the context-based knowledge information that has been demonstrated. The future work of this paper is to develop a decision making system based on knowledge reuse using the technique of Case Based Reasoning (CBR).